\documentclass[10pt,twocolumn,letterpaper]{article}

\usepackage[pagenumbers]{iccv} 

\usepackage[utf8]{inputenc}       
\usepackage[T1]{fontenc}          
\usepackage{lmodern}              
\usepackage{amsmath, amssymb, amsfonts, bm} 
\usepackage{graphicx}             
\usepackage{booktabs}             
\usepackage{multirow}             
\usepackage{float}                
\usepackage{url}                  
\usepackage{xcolor}               

\usepackage{caption}
\usepackage{subcaption}           
\usepackage{algorithm}            
\usepackage{algpseudocode}        
\usepackage{algorithmicx}
\floatname{algorithm}{Algorithm}

\usepackage[numbers,sort&compress]{natbib}  

\usepackage{geometry}
\usepackage{setspace}
\usepackage{siunitx}              
\usepackage{enumitem}             
\usepackage{microtype}            

\definecolor{iccvblue}{rgb}{0.21,0.49,0.74}
\usepackage[pagebackref,breaklinks,colorlinks,allcolors=iccvblue]{hyperref}

\def\paperID{*****} 
\def\confName{ICCV}
\def\confYear{2025}

\title{Motion-Aware Reinforcement Learning For Object Localization}

\author{%
  Prithvi Raj Singh\thanks{Corresponding author}\\
  McNeese State University\\
  {\tt\small psingh8@mcneese.edu}
  \and
  Satyendra Singh\\
  Louisiana Tech University\\
  {\tt\small srs083@latech.edu}
}

\begin{document}
\maketitle
\begin{abstract}
We present \textbf{MARLNet} (Motion-Aware Reinforcement Learning
Network), a PPO-based bounding-box refinement agent that incorporates
a constant-velocity motion prior into the observation state and an
action smoothness penalty into the reward function.
The agent operates on 268-dimensional observations encoding the
current proposal, a kinematic prediction, the previous action, and
a 256-dimensional EfficientNet-B0 crop feature, and learns a
five-dimensional policy controlling coordinate adjustments and a
binary termination trigger.
Evaluated on Pascal VOC 2012 and VisDrone 2019, MARLNet trains
stably across all regularization strengths tested and achieves
consistent gains in detection success rate at
$\text{IoU} \geq 0.5$: up to $+0.011$ on VOC
($\lambda_\text{phys}{=}0.10$), where the motion prior prevents
the overshooting that causes plain PPO to regress on this metric,
and $+0.007$ on VisDrone ($\lambda_\text{phys}{=}0.70$), where
unconstrained PPO achieves a larger gain ($+0.025$) owing to the
weaker base detector.
Through reward design ablations and training dynamics analysis, we
identify a reward interference in which combining a constant-velocity
deviation penalty with an absolute IoU term causes trigger collapse, and show that replacing it with the action smoothness
penalty resolves this failure.
We further characterize a representational ceiling facing crop-feature
refinement agents that share a backbone with their base detector,
confirmed through a global-plus-local observation ablation. 
\noindent\textit{Project page:}
\url{https://prithviraj97.github.io/marl-net}

\end{abstract}    
\section{Introduction}
\label{sec:intro}

Modern object detectors produce bounding-box predictions in a
single forward pass, mapping image features directly to box
coordinates. While this approach achieves strong accuracy on standard benchmarks, it
treats localization as a one-shot regression problem with no
opportunity for iterative correction.
A detector that correctly identifies an object can still produce a
proposal that is misaligned in position, scale, or aspect ratio --
errors that compound under strict overlap-based evaluation criteria
such as AP@.75 and are especially consequential for small or
densely packed objects.
Post-hoc refinement addresses this gap directly: rather than retraining
the detector, a lightweight module takes each committed proposal as
input and iteratively adjusts its coordinates toward the ground truth.
This decoupling makes refinement architecturally flexible and
computationally efficient, and it has motivated a growing line of
work seeking to close the localization gap without the cost of
full detector retraining~\cite{bbrefinement, refinebox, alpha-refine, zhou2021reinforcenet}.

Reinforcement learning (RL) offers a compelling approach to bounding-box refinement 
by treating localization as a sequence of corrective decisions rather than a single prediction.
Refinement is inherently sequential -- each coordinate adjustment
changes the crop seen by the next step -- and the target signal,
intersection-over-union, is non-differentiable and therefore
inaccessible to standard regression losses.
Policy gradient methods~\cite{ppoalgorithms} handle both properties directly.
Within this RL formulation, motion prior is an
appealing design component: a bounding box being refined should move smoothly, and a kinematic model encoding this expectation --
such as a constant-velocity prediction that extrapolates the box's
recent trajectory -- provides a principled inductive bias that could
regularize the policy and discourage erratic behavior~\cite{bewley2016simple, mysore2021regularizing}. Both components have independently attracted sustained research interest~\cite{caicedo2015active, zhou2021reinforcenet,
bewley2016simple, guo2024pmtrack}, and their combination in a
unified framework is a natural extension that, to our knowledge,
has not been systematically studied.

What is less well understood is how motion-derived penalty terms interact with the remaining components of a reward function designed
for IoU optimization. In practice, reward design for RL-based refinement involves at least
three coupled objectives: rewarding overlap improvement, penalizing
unnecessary steps, and incentivizing timely termination via a
learned trigger.
Adding a motion-derived term introduces a fourth, and the
interactions among all four are not transparent.
Equally underexplored is the question of representational capacity:
if the refinement agent operates on crop features extracted by the
same backbone as the base detector, what ceiling does that shared
basis impose on achievable performance, and is that ceiling a
consequence of the crop-only view or a deeper representational
constraint? These two questions motivate the present work.

We address both issues through the design and systematic evaluation
of \textbf{MARLNet} (Motion-Aware Reinforcement Learning Network),
a PPO-based bounding box refinement agent that incorporates a
constant-velocity motion prior into the observation and an action
smoothness penalty into the reward.
MARLNet represents, to our knowledge, the first RL-based refinement
framework to explicitly study the role of motion-structured priors
in the single-image iterative refinement setting.
In constructing and evaluating this framework across Pascal VOC
2012~\cite{voc} and VisDrone 2019~\cite{visdrone}, we uncover two
findings that carry implications beyond our specific implementation.
First, we identify a reward interference in which a constant-velocity
deviation penalty, combined with an absolute IoU term, causes trigger
probability collapse by training epoch~26 and reduces mAP from~0.490
to~0.035 on VOC (see appendix~\ref{sec:ablation}).
Replacing this with an action smoothness penalty eliminates the
collapse entirely. Second, we characterize a representational ceiling that bounds the performance of any crop-feature refinement agent sharing a backbone
with its base detector -- a constraint that is architectural, not
a consequence of reward design, and one that points toward the next
generation of refinement architectures.

Through controlled ablations over reward variants, motion penalty
strengths, and observation designs, we establish that this ceiling
is representational: the agent and detector share an EfficientNet-B0~\cite{efficientnet}
backbone, and no reformulation of the crop-only view -- including providing the agent with a full-image global feature alongside the
crop -- consistently overcomes this constraint.
The agent does achieve reliable gains in detection success rate at
$\text{IoU} \geq 0.5$ on both datasets, a metric that reflects
practical precision requirements in downstream applications.

\noindent Our contributions are as follows:
\begin{itemize}

    \item \textbf{MARLNet}, a motion-aware reinforcement learning
          framework for iterative bounding-box refinement.
          MARLNet integrates a constant-velocity motion prior into
          the observation state and an action smoothness penalty
          into the reward, enabling stable PPO training across all
          regularization strengths tested on Pascal VOC 2012 and VisDrone 2019.

    \item A characterization of the \emph{feature information
          ceiling} inherent to crop-feature RL refinement agents
          that share a backbone with their base detector.
          A global-plus-local observation ablation confirms the
          ceiling is representational rather than observational,
          providing a concrete architectural diagnosis that
          motivates future design choices.

    \item A reproducible experimental framework covering reward
          variant ablations, motion regularization sensitivity
          ($\lambda_{\text{phys}} \in \{0.05, 0.10, 0.20,
          0.50, 0.70\}$), latency benchmarking, and cross-dataset
          evaluation on two domains with substantially different
          object density, scale, and viewpoint.

    \item An analysis of reward interference in
          motion-regularized RL refinement, tracing training
          collapse to the combination of a kinematic deviation
          penalty and an absolute IoU term, with empirical
          evidence confirmed across both datasets.

\end{itemize}

\section{Related Work}
\label{sec:related}

\subsection{Bounding Box Refinement}

Improving the spatial precision of detector proposals has been a
long-standing objective in computer vision.
Classical approaches couple a lightweight regression head directly
to the detector backbone, refining coordinates in a single
forward pass~\cite{faster-rcnn}.
More recent work has extended this idea to transformer-based
architectures: RefineBox~\cite{refinebox} applies a frozen
backbone with a dedicated regression module to refine DETR-style
predictions, while BBRefinement~\cite{bbrefinement} proposes
a universal post-hoc scheme that improves localization precision
across multiple detector families without retraining.
In the visual tracking literature, Alpha-Refine~\cite{alpha-refine}
applies high-resolution refinement to coarse tracker predictions,
demonstrating that a dedicated refinement stage consistently improves
overlap with the target object.
Iterative refinement approaches push this further by applying
the correction step multiple times: Cruciata et al.~\cite{cruciata2022iterative}
show that iterating bounding-box refinement within a tracking pipeline
reduces accumulated drift compared to single-pass correction.
A shared limitation of all regression-based methods is that refinement
is performed without sequential decision-making -- the model commits
to a correction in one step and cannot revise that decision in light
of new evidence.
Reinforcement learning addresses this limitation directly by
framing refinement as a multi-step policy optimization problem.

\subsection{RL for Object Localization}

Caicedo and Lazebnik~\cite{caicedo2015active} established the
foundational MDP formulation for RL-based object localization,
training a DQN agent to iteratively deform a bounding box using
eight discrete transformation actions with an IoU-based reward
and a learned termination trigger.
Subsequent work extended this framework to multi-class
detection~\cite{mathe2016reinforcement}, proposal
generation~\cite{pirinen2018deep}, and post-hoc refinement
applied after a separate detector has committed to a
proposal~\cite{zhou2021reinforcenet}.
K\"{o}nig et al.~\cite{konig2019multi} explicitly study the effect
of reward design on convergence and final localization quality
across multiple training stages, finding that the choice of reward
signal and step penalty substantially affects both speed and
performance -- a finding directly relevant to our analysis
of reward interference.
In the tracking domain, RL agents have been trained to adjust
viewpoint and bounding box parameters in response to target
appearance~\cite{kurl2024}, and applied to drone imagery where
fast-moving targets require robust sequential localization
policies~\cite{ozer2021visual}.
Despite this breadth of application, the interaction between
motion-derived regularization terms and IoU-based reward
components has not been studied in any of these settings --
a gap the present work addresses.

\subsection{Motion Priors in Object Tracking and Detection}
Motion priors are well-established in multi-object tracking,
where the constant-velocity Kalman filter provides a principled
temporal state prior that extrapolates track positions from one
video frame to the next~\cite{bewley2016simple}.
The appeal is grounded in the dynamics of real scenes:
objects move smoothly, and encoding this expectation reduces
sensitivity to missed detections and noisy proposals.
Recent work has replaced hand-designed Kalman dynamics with
learned motion models~\cite{advzemovic2025beyond}, improving
tracking robustness under non-linear motion and partial
occlusion.
In the detection literature, motion cues have been incorporated
through attention mechanisms to improve frame-to-frame
consistency in traffic surveillance~\cite{liu2023motionprior},
while PMTrack~\cite{guo2024pmtrack} introduces motion-aware
multi-object association that compensates for camera shake and
non-linear object trajectories.
MotionTrack~\cite{xiao2024motiontrack} employs a learnable
motion predictor to forecast object positions in dynamic scenes,
demonstrating that learned priors generalize better than
hand-designed kinematic models when motion patterns are
irregular. Kinematic constraints have also proven effective in 3D detection:
Brazil et al.~\cite{brazil2020kinematic} show that incorporating
motion consistency into monocular video detection improves
temporal stability under occlusion.

Despite this body of evidence, all of the above methods operate
across video frames, where the constant-velocity assumption
reflects genuine physical object motion. The single-image iterative refinement setting studied here is
categorically different -- the agent acts within one static
image, and the sequential steps carry no genuine temporal
dynamics. Whether and how motion priors transfer to this setting
is an open question that our present work addresses.

\subsection{Action Smoothness in RL}
Independent of the detection literature, a body of work in
continuous control has studied the effect of action smoothness
regularization on policy stability and generalization.
Mysore et al.~\cite{mysore2021regularizing} introduced
conditioning for action policy smoothness (CAPS), which
penalizes the $\ell_2$ norm of consecutive action differences
during policy optimization and demonstrated improved training
stability and sim-to-real transfer in robotic control tasks.
Lee et al.~\cite{lee2024gradient} extended this principle to
gradient-based smoothness constraints, showing that penalizing
the gradient of the policy with respect to the state reduces
policy sensitivity to observation noise.
A consistent finding across this line of work is that excessive
smoothing reduces policy expressiveness -- a phenomenon
consistent with the non-monotonic relationship between the
smoothness coefficient $\lambda$ and generalization performance
that we observe in Section~\ref{sec:experiments}.
In the model-based RL literature, learning an explicit predictive
dynamics model has been shown to improve policy stability and
sample efficiency when the model operates in a low-dimensional
state space~\cite{ebert2018visualforesight}, providing a
theoretical precedent for incorporating structured priors into
the reward computation.
To our knowledge, action smoothness penalties have not previously
been applied to RL-based bounding box refinement, nor has their
interaction with IoU-based detection rewards been studied in any
published work.

\section{Problem Formulation}
\label{problem_formulation}

We formulate iterative bounding box refinement as a finite-horizon Markov Decision Process (MDP), defined by the tuple $\mathcal{M} = (\mathcal{S}, \mathcal{A}, \mathcal{P}, \mathcal{R}, \gamma, T)$. Rather than predicting box coordinates in a single forward pass, the agent learns to refine an initial detection hypothesis through a sequence of geometry-adjusting actions, guided by a reward signal that encodes both localization accuracy and trajectory smoothness.

\subsection{Markov Decision Process Formulation}

\paragraph{State Space $\mathcal{S}$.}
At each time step $t \in \{0, 1, \ldots, T\}$, the agent observes a state vector
\begin{equation}
\label{eq_state}
s_t = \bigl[b_t,\; \hat{b}^{\text{cv}}_t,\; a_{t-1},\; f_t\bigr] \in \mathcal{S},
\end{equation}
where:
\begin{itemize}
    \item $b_t = (x_t, y_t, w_t, h_t) \in \mathbb{R}^4$ is the current bounding box, with coordinates normalized to $[0,1]$.
    \item $\hat{b}^{\text{cv}}_t \in \mathbb{R}^4$ is the constant-velocity prediction of the next box position.
    \item $a_{t-1} = (\delta x_{t-1}, \delta y_{t-1}, \delta w_{t-1}, \delta h_{t-1}) \in \mathbb{R}^4$ is the previous action (initialized to zero at $t=0$).
    \item $f_t \in \mathbb{R}^{256}$ is a visual feature embedding
      extracted from the image crop $\mathcal{I}(b_t)$, where $\mathcal{I}(b_t)$ denotes the region of the image bounded
      by $b_t$, resized to $224 \times 224$ pixels before feature extraction.
\end{itemize}

The concatenated state vector is 268-dimensional ($4 + 4 + 4 + 256$).
For the global-plus-local observation discussed in
(~\ref{sec:ceiling_discussion}), a second full-image
EfficientNet-B0~\cite{efficientnet} embedding is appended alongside $f_t$,
yielding a 524-dimensional state; this variant is evaluated
separately and not used for the main reported results.

The initial box $b_0$ is provided by a pre-trained object detector.

\paragraph{Action Space $\mathcal{A}$.}
The agent produces a continuous action
$a_t = (\delta x_t, \delta y_t, \delta w_t, \delta h_t) \in \mathbb{R}^4$
and a binary stopping signal $a^{\text{stop}}_t \in \{0,1\}$.
Before being applied to the environment, continuous components
are clamped to $[-\delta_{\max}, \delta_{\max}]$ with
$\delta_{\max} = 0.05$.
The box is then updated as:
\begin{align}
c^x_{t+1} &= \operatorname{clamp}(c^x_t + \delta x_t \cdot w_t), \\
c^y_{t+1} &= \operatorname{clamp}(c^y_t + \delta y_t \cdot h_t), \\
w_{t+1}   &= \operatorname{clamp}\bigl(w_t(1 + \delta w_t)\bigr), \\
h_{t+1}   &= \operatorname{clamp}\bigl(h_t(1 + \delta h_t)\bigr),
\end{align}
where $(c^x_t, c^y_t)$ denotes the box centre.
All resulting box coordinates are clamped to $[0,1]$ to keep
the proposal within the image boundary.
Raw pre-clamp action values are stored in the rollout buffer
so that PPO log-probability computations remain consistent with
the sampling distribution.

\paragraph{Transition Dynamics $\mathcal{P}$.}
The transition function $\mathcal{P}(s_{t+1} \mid s_t, a_t)$ is deterministic. The next state is formed by updating the box, computing the new constant-velocity prediction, storing $a_t$ as the previous action, and extracting features from $\mathcal{I}(b_{t+1})$.

\paragraph{Reward Function $\mathcal{R}$.}
The agent receives a scalar reward:
\begin{equation}
r_t = \alpha \cdot \Delta \text{IoU}_t - c_{\text{step}} - \lambda_{\text{phys}} \cdot \|a_t - a_{t-1}\|_2,
\end{equation}
where
\[
\Delta \text{IoU}_t = \text{IoU}(b_{t+1}, b^*) - \text{IoU}(b_t, b^*).
\]
The smoothness penalty is not applied at $t=0$ since no previous
action exists; $\lambda_\text{phys}$ acts from $t=1$ onward. At termination, the agent receives:
\begin{equation}
r^{\text{terminal}} = C_{\text{bonus}} \cdot \text{IoU}(b_{t+1}, b^*).
\end{equation}

Termination occurs under three conditions.
If the stopping trigger fires ($a^\text{stop}_t > 0.5$) or the
refined box reaches $\text{IoU}(b_{t+1}, b^*) \geq \tau_\text{stop}$,
the episode ends and the agent receives a terminal bonus proportional
to final localization quality:
\begin{equation}
r^\text{terminal} = C_\text{bonus} \cdot \text{IoU}(b_{t+1}, b^*),
\quad C_\text{bonus} = 5.0.
\end{equation}
If neither condition is met and $t = T$, the episode ends with no
terminal bonus. No penalty is applied when the trigger fires before
reaching $\tau_\text{stop}$; the per-step cost $c_\text{step}$
naturally discourages unnecessary steps without destabilising the
trigger. We use $\tau_\text{stop} = 0.6$ for VOC and $\tau_\text{stop} = 0.5$
for VisDrone, $\gamma = 0.99$, and $T = 15$.

\subsection{Motion Prior}
\label{physics_prior}
We use a constant-velocity (CV) model:
\begin{equation}
\hat{b}^{\text{cv}}_{t+1} = b_t + (b_t - b_{t-1}).
\end{equation}

This prediction is included in the state as a momentum cue. It is not used in the reward; instead, smoothness is enforced via the action penalty $\|a_t - a_{t-1}\|_2$.
At $t=0$, no history exists; the model returns $\hat{b}^\text{cv}_1
= b_0$, representing a zero predicted displacement.

\subsection{Learning Objective}

We learn a policy $\pi_\theta(a_t \mid s_t)$ maximizing:
\begin{equation}
J(\theta) = \mathbb{E}_{\tau \sim \pi_\theta} \left[ \sum_{t=0}^{T} \gamma^t r_t \right].
\end{equation}

We optimize using PPO with objective:
\begin{multline}
\mathcal{L}^{\text{PPO}}(\theta) = \mathbb{E}_t \Big[ \min \big( r_t(\theta)\hat{A}_t, \\
\operatorname{clip}(r_t(\theta), 1-\epsilon, 1+\epsilon)\hat{A}_t \big) \Big]
\end{multline}
where
\[
r_t(\theta) = \frac{\pi_\theta(a_t \mid s_t)}{\pi_{\theta_{\text{old}}}(a_t \mid s_t)},
\quad
\hat{A}_t = \hat{R}_t - V_\phi(s_t).
\]

\section{Methodology}
\label{sec:methodology}

\begin{figure*}[t]
    \centering
    \includegraphics[width=\textwidth]{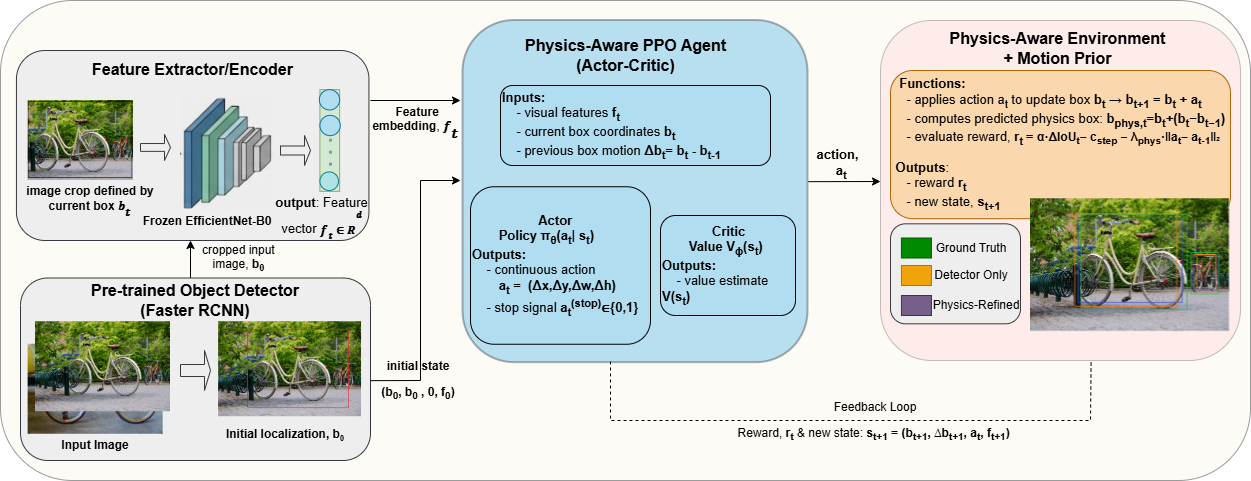}
    \caption{
    Overview of the proposed \textbf{motion-aware reinforcement learning framework} for bounding box refinement.
    Given an initial object hypothesis from a detector, visual features are extracted from the cropped region and passed to an actor–critic agent.
    The agent predicts a bounding box adjustment and a stop signal.
    The environment evaluates the resulting IoU improvement and applies a motion-based consistency penalty derived from a constant-velocity model.
    The reward guides the PPO agent toward accurate and motion-consistent localization.
    }
    \label{ppo_arch}
\end{figure*}

This section describes the architecture and training procedure of
MARLNet in full detail. The framework comprises three tightly coupled
components: (1)~a frozen visual feature extractor that provides a
stable, appearance-based state representation; (2)~a motion-aware
refinement environment that manages bounding-box transitions and
computes rewards; and (3)~a PPO-based actor-critic agent that learns
to balance localization accuracy with trajectory smoothness.
Figure~\ref{ppo_arch} provides a schematic overview of the
complete pipeline.

\subsection{System Architecture}
\label{sec:architecture}

\subsubsection{Initial Detection}

Given an input RGB image $\mathcal{I} \in \mathbb{R}^{H \times W \times 3}$,
an initial bounding-box hypothesis $b_0 = (x_0, y_0, w_0, h_0)$ is
obtained from a pre-trained object detector, specifically a
Faster R-CNN~\cite{faster-rcnn} model with a ResNet-50~\cite{he2016deep} backbone.
The detector serves as a fixed initialization module; its weights are
not updated during RL training.  All coordinates are normalized to
$[0,1]$ relative to the image dimensions, ensuring scale invariance
across images of different resolutions.

\subsubsection{Visual Feature Extraction}
\label{sec:features}

At each time step $t$, the current bounding box $b_t$ defines a
region of interest within $\mathcal{I}$.  We crop this region,
resize it to $224 \times 224$ pixels, and pass it through a
pre-trained EfficientNet-B0~\cite{efficientnet} backbone to
obtain a visual embedding.  The global average pooling output of
EfficientNet-B0 (dimension 1280) is projected by a learned linear
layer to a compact feature vector $f_t \in \mathbb{R}^{256}$.

The backbone is frozen throughout RL training for two reasons.
First, simultaneous updates to the feature extractor and the RL
policy would introduce non-stationarity into the state representation,
which is known to destabilize value estimation and policy gradients.
Second, freezing the backbone eliminates backpropagation through the
convolutional network, substantially reducing memory usage and
per-step computation time.

\subsubsection{Actor-Critic Network Architecture}
\label{sec:actor_critic}

The RL agent follows the actor-critic paradigm.  The state vector
$s_t = [b_t,\, \hat{b}^{\mathrm{cv}}_t,\, a_{t-1},\, f_t]$
(as defined in Equation~\ref{eq_state}) is concatenated into a
single vector of dimension $d_s = 268$ and processed by a shared
two-layer MLP:

\begin{equation}
  h_t = \mathrm{ReLU}\!\left(W_2 \cdot \mathrm{ReLU}(W_1 \cdot s_t +
  b_1) + b_2\right),
  \label{eq:mlp}
\end{equation}

\noindent where $W_1 \in \mathbb{R}^{256 \times 268}$,
$W_2 \in \mathbb{R}^{256 \times 256}$, and $b_1, b_2$ are bias
vectors.  The hidden representation $h_t \in \mathbb{R}^{256}$ is
shared across the actor and critic heads.

\paragraph{Actor head.}
The actor outputs parameters for a factored action distribution.
For the continuous bounding-box adjustment $a_t \in \mathbb{R}^4$,
it produces a mean vector and a vector of log-standard-deviations:

\begin{align}
  \mu_t &= W_\mu h_t + b_\mu, \quad \mu_t \in \mathbb{R}^4,
  \label{eq:actor_mean} \\
  \log \sigma_t &= \ell_{\log\sigma},
  \label{eq:actor_std}
\end{align}

\noindent where $\ell_{\log\sigma} \in \mathbb{R}^4$ is a
globally-shared learnable parameter vector (not conditioned on
$h_t$), following standard practice in PPO implementations
\cite{ppoalgorithms}.  Actions are sampled as
$a_t \sim \mathcal{N}(\mu_t, \mathrm{diag}(\sigma_t^2))$.
To prevent excessively large updates, sampled actions are clamped
to $[-\delta_{\max}, \delta_{\max}]$ with $\delta_{\max} = 0.05$
before being applied to the environment; the unclamped samples are
stored in the replay buffer so that log-probability recomputation
during the PPO update remains exact.

The actor also outputs a stopping probability via a scalar logit:

\begin{equation}
  p_{\mathrm{stop}} = \sigma\!\left(w_{\mathrm{stop}}^{\top}
  h_t + b_{\mathrm{stop}}\right),
  \label{eq:stop}
\end{equation}

\noindent from which the binary stopping decision is sampled as
$a^{\mathrm{stop}}_t \sim \mathrm{Bernoulli}(p_{\mathrm{stop}})$.

\paragraph{Critic head.}
The critic estimates the scalar state-value function:

\begin{equation}
  V_\phi(s_t) = w_v^{\top} h_t + b_v,
  \label{eq:critic}
\end{equation}

\noindent which approximates the expected discounted return from
$s_t$ under the current policy and provides the baseline for
advantage computation.

\subsection{Motion-Aware Environment}
\label{sec:environment}

The environment maintains bounding-box state, computes state
transitions, generates the CV-prediction feature, and evaluates
the reward signal at each step.

\subsubsection{State Transitions}
\label{sec:transitions}

Upon receiving action $a_t = (\delta x_t, \delta y_t, \delta w_t,
\delta h_t)$, the environment updates the bounding box via a center-relative, scale-multiplicative rule.  Let $(c^x_t, c^y_t)$
denote the box center. Then:

\begin{align}
  c^x_{t+1} &= \mathrm{clamp}(c^x_t + \delta x_t \cdot w_t),
  \label{eq:cx} \\
  c^y_{t+1} &= \mathrm{clamp}(c^y_t + \delta y_t \cdot h_t),
  \label{eq:cy} \\
  w_{t+1}   &= \mathrm{clamp}(w_t(1 + \delta w_t)),
  \label{eq:w} \\
  h_{t+1}   &= \mathrm{clamp}(h_t(1 + \delta h_t)),
  \label{eq:h}
\end{align}

\noindent where all values are clamped to $[0,1]$ with the
additional constraint that the box remains within image
boundaries.  This parameterization is preferable to a purely
additive update $b_{t+1} = b_t + a_t$ because a fixed
$\delta x$ produces a displacement proportional to the current
box width, making the action magnitude invariant to object scale.
The new state $s_{t+1}$ is then constructed as described in
Section~\ref{problem_formulation}.

\subsubsection{Reward Computation}
\label{sec:reward}

The environment evaluates the quality of $b_{t+1}$ and returns a
scalar reward.  The complete reward at step $t$ is:

\begin{equation}
  r_t = \underbrace{\alpha \cdot \Delta\mathrm{IoU}_t}_{%
    \text{improvement}} \;-\;
  \underbrace{c_{\mathrm{step}}}_{\text{step cost}} \;-\;
  \underbrace{\lambda_{\mathrm{phys}} \cdot
    \|a_t - a_{t-1}\|_2}_{\text{smoothness penalty}},
  \label{eq:reward}
\end{equation}

\noindent where $\Delta\mathrm{IoU}_t =
\mathrm{IoU}(b_{t+1}, b^*) - \mathrm{IoU}(b_t, b^*)$ and
$b^*$ is the ground-truth box.  The IoU between two boxes is
computed as:

\begin{equation}
  \mathrm{IoU}(b, b^*) =
  \frac{\mathrm{Area}(b \cap b^*)}{\mathrm{Area}(b \cup b^*)}.
  \label{eq:iou}
\end{equation}

We denote the motion smoothness weight as $\lambda_{\text{phys}}$
throughout to distinguish it from the GAE parameter
$\lambda_{\text{GAE}}$; the subscript reflects the smoothness
penalty's role as a motion-consistency regularizer.

The three terms in Equation~\ref{eq:reward} each serve a distinct
purpose.  The improvement term $\alpha \cdot \Delta\mathrm{IoU}_t$
is the primary localization signal: it rewards steps that bring the
box closer to the ground truth and penalizes steps that move it
away.  The step cost $c_{\mathrm{step}}$ makes each refinement
step carry a small fixed cost, discouraging the agent from taking
unnecessary steps once a satisfactory localization has been
reached.  The smoothness penalty $\lambda_{\mathrm{phys}}
\|a_t - a_{t-1}\|_2$ penalizes abrupt changes in the action
direction across consecutive steps, encoding the inductive bias
that physically plausible refinement trajectories are smooth and
directionally consistent.  As discussed in
Section~\ref{physics_prior}, this penalty acts on the
action sequence rather than on the deviation from a predicted box
position; this distinction is critical to avoiding the reward interference
that arises from the constant-velocity reward formulation.

At episode termination -- triggered either by $a^{\mathrm{stop}}_t
= 1$ or by $\mathrm{IoU}(b_{t+1}, b^*) \geq \tau_{\mathrm{stop}}$
-- the agent receives an additional terminal bonus:

\begin{equation}
  r^{\mathrm{terminal}} = C_{\mathrm{bonus}} \cdot
  \mathrm{IoU}(b_{t+1}, b^*), C_{\mathrm{bonus}} = 5.0
  \label{eq:terminal}
\end{equation}
When the episode ends due to reaching $t = T_{\max}$, no
terminal bonus is awarded; only trigger-fired or
$\tau_\text{stop}$-triggered termination earns
$r^\text{terminal}$.
The proportional form of equation~\ref{eq:terminal} provides a
graded stopping incentive: the agent earns a larger bonus by
terminating at a well-aligned box than by stopping prematurely.
At $t = 0$, where no previous action exists, the smoothness
penalty is not applied.

\subsection{Training Procedure}
\label{sec:training}
The complete training procedure is summarized in Algorithm~\ref{alg:parlnet}.
\subsubsection{Experience Collection}

Training proceeds in epochs.  Each epoch collects a fixed buffer
of $N$ environment transitions by rolling out the current policy
$\pi_\theta$ across randomly sampled training images.  For each
episode, the environment is initialized with the detector output
$b_0$ and the state vector $s_0 = [b_0,\, b_0,\, \mathbf{0}_4,\,
f_0]$, where the second $b_0$ serves as the initial CV prediction
(since no prior box exists at $t=0$) and $\mathbf{0}_4$ initializes
the previous-action slot.  Each transition
$(s_t, a_t, r_t, \log\pi_\theta(a_t|s_t), V_\phi(s_t))$ is stored
in a replay buffer $\mathcal{B}$.  An episode terminates when
$a^{\mathrm{stop}}_t = 1$, when $\mathrm{IoU}(b_{t+1}, b^*)
\geq \tau_{\mathrm{stop}}$, or when $t = T_{\max}$.

\subsubsection{Advantage Estimation}

After collecting $\mathcal{B}$, advantage estimates are computed
using Generalized Advantage Estimation (GAE)~\cite{GAE}:

\begin{equation}
  \hat{A}_t = \sum_{l=0}^{T-t} (\gamma\lambda_{\mathrm{GAE}})^l
  \,\delta_{t+l},
  \label{eq:gae}
\end{equation}

\noindent where $\delta_t = r_t + \gamma V_\phi(s_{t+1}) -
V_\phi(s_t)$ is the temporal-difference error, $\gamma$ is the
discount factor, and $\lambda_{\mathrm{GAE}} \in [0,1]$ trades off
bias against variance in the advantage estimate.  Target returns
are computed as $\hat{R}_t = \hat{A}_t + V_\phi(s_t)$.

\subsubsection{Policy and Value Updates}

The actor is updated by maximizing the clipped PPO
objective~\cite{ppoalgorithms}:

\begin{multline}
\mathcal{L}^{\text{PPO}}(\theta) = \mathbb{E}_t \Big[ \min \big( r_t(\theta)\hat{A}_t, \\
\operatorname{clip}(r_t(\theta), 1-\epsilon, 1+\epsilon)\hat{A}_t \big) \Big]
\end{multline}

\noindent where $r_t(\theta) = \pi_\theta(a_t|s_t) /
\pi_{\theta_{\mathrm{old}}}(a_t|s_t)$ is the importance-sampling
ratio.  The critic is updated by minimizing the mean squared error
between predicted and target values:

\begin{equation}
  \mathcal{L}^{\mathrm{value}}(\phi) = \mathbb{E}_t\!\left[
  \bigl(V_\phi(s_t) - \hat{R}_t\bigr)^2\right].
  \label{eq:value_loss}
\end{equation}

To encourage exploration of the action and stopping distributions,
an entropy regularization term is added:

\begin{equation}
  \mathcal{L}^{\mathrm{entropy}} = -c_e\,\mathbb{E}_t
  \bigl[\mathcal{H}[\pi_\theta(\cdot|s_t)]\bigr],
  \label{eq:entropy}
\end{equation}

\noindent where $\mathcal{H}[\cdot]$ denotes Shannon entropy and
$c_e$ is the entropy coefficient.  The combined training loss is:

\begin{equation}
  \mathcal{L}^{\mathrm{total}} = \mathcal{L}^{\mathrm{PPO}}(\theta)
  + c_v\,\mathcal{L}^{\mathrm{value}}(\phi)
  + \mathcal{L}^{\mathrm{entropy}},
  \label{eq:total_loss}
\end{equation}

\noindent where $c_v$ is the value loss coefficient.  For each
collected buffer, the network is updated for $K$ epochs of
minibatch gradient descent using the Adam optimizer~\cite{kingma2017adam}, after which a new buffer is
collected with the updated policy.  Gradient norms are clipped to $0.5$ to prevent destabilizing updates.

Detailed implementation settings are provided in the Appendix~\ref{sec:impl} for reproducibility, including model architecture, 
feature extraction configuration, optimization hyperparameters, and hardware specifications.


\begin{algorithm}[t]
\caption{MARLNet Training with Motion-Aware PPO}
\label{alg:parlnet}
\begin{algorithmic}[1]
\Require Pre-trained detector $\mathcal{D}$, training dataset
         $\mathcal{D}_{\mathrm{train}}$, frozen feature extractor
         $\Phi$, hyperparameters
\State Initialize policy $\pi_\theta$, value network $V_\phi$
\State Initialize motion-aware environment $\mathcal{E}$
\For{each training epoch}
  \State $\mathcal{B} \leftarrow \emptyset$
         \Comment{Initialize replay buffer}
  \For{each sampled image $\mathcal{I}$ from $\mathcal{D}_{\mathrm{train}}$}
    \State $b_0 \leftarrow \mathcal{D}(\mathcal{I})$
           \Comment{Initial box from detector}
    \State $f_0 \leftarrow \Phi(\mathcal{I}(b_0))$
           \Comment{Extract visual features}
    \State $s_0 \leftarrow [b_0,\; b_0,\; \mathbf{0}_4,\; f_0]$
           \Comment{$\hat{b}^{\mathrm{cv}}_0 = b_0$; $a_{-1} = \mathbf{0}$}
    \For{$t = 0$ to $T_{\max}$}
      \State Sample $a_t \sim \pi_\theta(a_t | s_t)$,\;
             $a^{\mathrm{stop}}_t \sim \mathrm{Bernoulli}(p_{\mathrm{stop}})$
      \State Execute $a_t$ in $\mathcal{E}$, observe $r_t$, $s_{t+1}$
      \State Store $(s_t, a_t, r_t,
             \log\pi_\theta(a_t|s_t), V_\phi(s_t))$ in $\mathcal{B}$
      \If{$a^{\mathrm{stop}}_t = 1$ \textbf{or}
          $\mathrm{IoU}(b_{t+1}, b^*) \geq \tau_{\mathrm{stop}}$
          \textbf{or} $t = T_{\max}$}
        \State \textbf{break}
      \EndIf
    \EndFor
  \EndFor
  \State Compute advantages $\{\hat{A}_t\}$ via GAE on $\mathcal{B}$
  \State Compute target returns $\{\hat{R}_t\}$
  \For{$k = 1$ to $K$ update epochs}
    \State Shuffle $\mathcal{B}$ and partition into minibatches
    \For{each minibatch}
      \State Compute $\mathcal{L}^{\mathrm{total}}$
             (Eq.~\ref{eq:total_loss})
      \State Update $\theta$, $\phi$ via Adam; clip gradients to $0.5$
    \EndFor
  \EndFor
\EndFor
\end{algorithmic}
\end{algorithm}

\section{Experimental Details}
\label{sec:experiments}

\subsection{Datasets}
\label{sec:datasets}

\paragraph{Pascal VOC2012.}
We train and evaluate primarily on the Pascal VOC2012
dataset~\cite{voc}, which contains 11,540 images
with bounding-box annotations across 20 object categories.
We use the standard \texttt{train} split (5,717 images) for
policy training and the \texttt{val} split (5,823 images) for
evaluation. Initial bounding-box hypotheses are generated by a
pre-trained Faster R-CNN~\cite{faster-rcnn} with a ResNet-50
backbone, run at a confidence threshold of $0.5$.

\paragraph{VisDrone2019-DET.}
To assess generalization to a more challenging domain, we evaluate MARLNet on the VisDrone2019 detection
benchmark~\cite{visdrone}.  VisDrone features aerial imagery from UAV platforms with densely packed, small-scale
objects under significant viewpoint and illumination variation. Results on VisDrone are reported in Section~\ref{sec:visdrone}.

COCO-pretrained Faster R-CNN cannot be used directly on VisDrone because the object classes 
are incompatible (80 COCO classes vs 10 VisDrone categories) and the domain 
shift from ground-level to aerial imagery is severe enough to produce near-random 
initial detections. We fine-tuned a ResNet-50 Faster R-CNN on the VisDrone2019 training 
split for 30 epochs, keeping the backbone pretrained on ImageNet and replacing the 
classification head. Training converged around epoch 15 with loss plateauing at 
approximately 0.85. The resulting detector achieves mAP 0.189 on the VisDrone2019- 
validation split, which is consistent with published results~\cite{du2019visdrone, zhao2025lpae, cao2024visible} for this architecture 
on this dataset.

\noindent Ground-truth annotations serve exclusively as reward targets during training and as evaluation references at test time; they are never
provided to the agent as input.

\begin{figure*}[!htp]
    \centering
    \includegraphics[width=\textwidth]{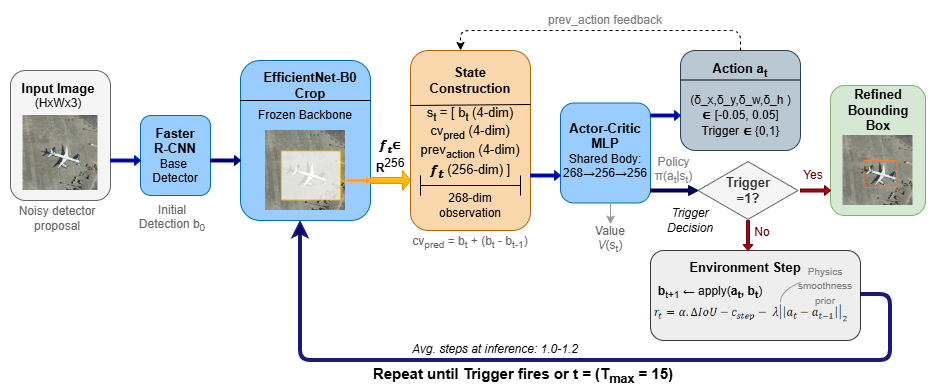}
    \caption{Inference pipeline of MARLNet. Given an input image, a
    fine-tuned Faster R-CNN provides an initial bounding-box
    hypothesis. At each refinement step, the current box defines
    a crop that is processed by a frozen EfficientNet-B0 backbone
    to produce a 256-dimensional feature vector $f_t$. This is
    concatenated with the current box coordinates, a
    constant-velocity motion prediction, and the previous action
    to form the 268-dimensional state $s_t$, which is passed to a
    shared actor-critic MLP. The actor outputs a continuous
    adjustment $({\delta x}, {\delta y}, {\delta w}, {\delta h})$
    and a binary stopping signal. If the trigger fires, the
    current box is returned as the refined output; otherwise the
    box is updated and the loop repeats. The motion smoothness penalty
    $\lambda\|a_t - a_{t-1}\|$ in the per-step reward encourages
    directionally consistent refinement trajectories.
    Maximum episode length $T_{\max} = 15$; average steps at
    inference: 1.0--1.2.}
    \label{fig:qualitative}
\end{figure*}

\subsection{Baselines}
\label{sec:baselines}

We compare MARLNet against four configurations that isolate the
contribution of each design component:

\begin{itemize}
  \item \textbf{Detector Only.}  Direct output of the pre-trained
    Faster R-CNN detector, without any further refinement.  This
    baseline establishes the quality of the initial hypothesis that
    all refinement methods start from.

  \item \textbf{Heuristic Policy.}  A hand-designed refinement
    rule that greedily shifts the box center and adjusts its scale
    toward the ground-truth location using IoU-based updates at
    each step.  It uses no learned components and serves as a
    lower bound on what a principled sequential strategy should
    achieve. We describe the heuristic policy and outline the refinement algorithm in detail in Appendix~\ref{app:heuristic}.

  \item \textbf{PPO (No Motion Priors).}  A PPO agent trained with
    $\lambda_{\mathrm{phys}} = 0$, receiving only the IoU
    improvement signal $\alpha \cdot \Delta\mathrm{IoU}_t -
    c_{\mathrm{step}}$ and the terminal bonus.  This isolates the
    contribution of the smoothness regularizer from the RL
    training itself.

  \item \textbf{MARLNet.}  The full model trained with the
    smoothness (motion) penalty ($\lambda_{\mathrm{phys}} = 0.05$) as
    described in Section~\ref{sec:reward}.  
\end{itemize}
Unless otherwise
    stated, results use the epoch-70 checkpoint, which we identify
    as the best-generalizing model based on validation performance
    (discussed further in Section~\ref{sec:training_dynamics}).

\subsection{Evaluation Metrics}
\label{sec:metrics}

We report a comprehensive set of localization metrics evaluated
on the VOC \texttt{val} split using the COCO evaluation
protocol~\cite{lin2014microsoft}:

\begin{itemize}
  \item \textbf{mAP (0.5:0.95).}  Mean Average Precision averaged
    over IoU thresholds from 0.50 to 0.95 in steps of 0.05.  This
    is the primary detection quality metric and is sensitive to
    both localization correctness and precision.

  \item \textbf{AP@0.5, AP@0.75.}  Average Precision at specific
    IoU thresholds.  AP@0.5 measures loose positional agreement;
    AP@0.75 measures tight precision.  Comparing the two reveals
    whether a method achieves correct localization but insufficient
    precision.

  \item \textbf{AR.}  Average Recall at 100 detections per image.

  \item \textbf{Mean IoU (mIoU).}  Best-match IoU between each predicted
    box and all ground-truth boxes for the same image, averaged
    across all predictions.

  \item \textbf{Success Rate (SR@$\tau$).}  Fraction of predicted
    boxes whose best-match IoU exceeds threshold $\tau$, reported
    at $\tau \in \{0.5, 0.7\}$.

  \item \textbf{$\Delta$IoU.}  Mean change in best-match IoU
    relative to the detector initialization, measuring net
    refinement gain or loss.
\end{itemize}

\section{Results \& Discussion}
\label{sec:discussion}

\subsection{Results on Pascal VOC 2012}
\label{sec:voc_discussion}

Table~\ref{tab:main_results} reports quantitative results for all four
methods on the Pascal VOC 2012 validation split. Table~\ref{tab:voc_comparison} presents a per-category breakdown 
of AP@0.5 for all methods. The Faster R-CNN base detector achieves mAP~0.490, AP@.5~0.748,
and AR~0.593, establishing a strong baseline against which all refinement methods are compared.

No refinement method improves mAP over the detector.
PPO without motion prior attains mAP~0.452 and AP@.5~0.745,
recovering most of the detector's performance but not surpassing it.
MARLNet ($\lambda{=}0.10$) achieves mAP~0.437 and AP@.5~0.750,
slightly below PPO on mAP while maintaining a comparable AP@.5.
The heuristic baseline produces the weakest mAP overall
(0.414, AP@.5~0.661), confirming that naive coordinate
adjustment without learned policy or motion-aware structure is
insufficient for reliable refinement.
All three refinement methods produce negative mean $\Delta$IoU
relative to the detector proposals, indicating that starting
from an already strong detection baseline, the refinement loop
tends on average to slightly perturb rather than improve
individual box coordinates.

The SR@.5 metric reveals a qualitatively different pattern.
Plain PPO achieves SR@.5~0.501, marginally below the detector's 0.504 -- indicating that
an unconstrained policy does not reliably improve precision
under strict IoU thresholds on this dataset. MARLNet with $\lambda{=}0.10$, by contrast, achieves
SR@.5~0.515, exceeding the detector by $+0.011$. This distinction is directionally consistent with the role
of the smoothness penalty: by constraining the magnitude of
consecutive action changes, the motion prior reduces the
probability of overshooting proposals that are already
well-localized, preserving precision where plain PPO's
unconstrained updates slightly degrade it. The SR@.7 metric follows a similar pattern, with PPO
recovering to 0.377 and MARLNet to 0.383, both below the detector's 0.389.

The method ranking on all mAP-based metrics is consistent
with the detector representing a performance ceiling:
Detector $>$ PPO $>$ MARLNet $>$ Heuristic.
We analyze the source of this ceiling in Section~\ref{sec:ceiling_discussion}.

\begin{table*}[!ht]
  \centering
  \begin{tabular}{lccccccc}
    \toprule
    Method & mAP & AP@.5 & AP@.75 & AR & SR@.5 & SR@.7 & $\Delta$IoU \\
    \midrule
    Detector only       & \textbf{0.490} & 0.748 & \textbf{0.544}
                        & \textbf{0.593} & 0.504 & \textbf{0.389} & --- \\
    \midrule
    Heuristic           & 0.414 & 0.661 & 0.442 & 0.557 & 0.506 & 0.359 & $-$0.013 \\
    PPO    & 0.452 & 0.745 & 0.498 & 0.555 & 0.501 & 0.377 & $-$0.012 \\
    MARLNet ($\lambda{=}0.10$) & 0.437 & \textbf{0.750} & 0.491
                         & 0.544 & \textbf{0.515}$^\uparrow$
                         & 0.383 & $-$0.008 \\
    \bottomrule
  \end{tabular}
  \caption{Localization results on Pascal VOC 2012 validation split (5823 images).
    All methods use the same Faster R-CNN initial detections.
    MARLNet uses the epoch-70 checkpoint with $\lambda{=}0.10$.
    $\Delta$IoU measures mean IoU change relative to the initial
    detector proposal; negative values indicate the refinement
    does not improve average overlap over the detector.
    Best result per column in \textbf{bold}.}
  \label{tab:main_results}
\end{table*}

\begin{figure*}[!ht]
    \centering
    \includegraphics[width=0.95\textwidth]{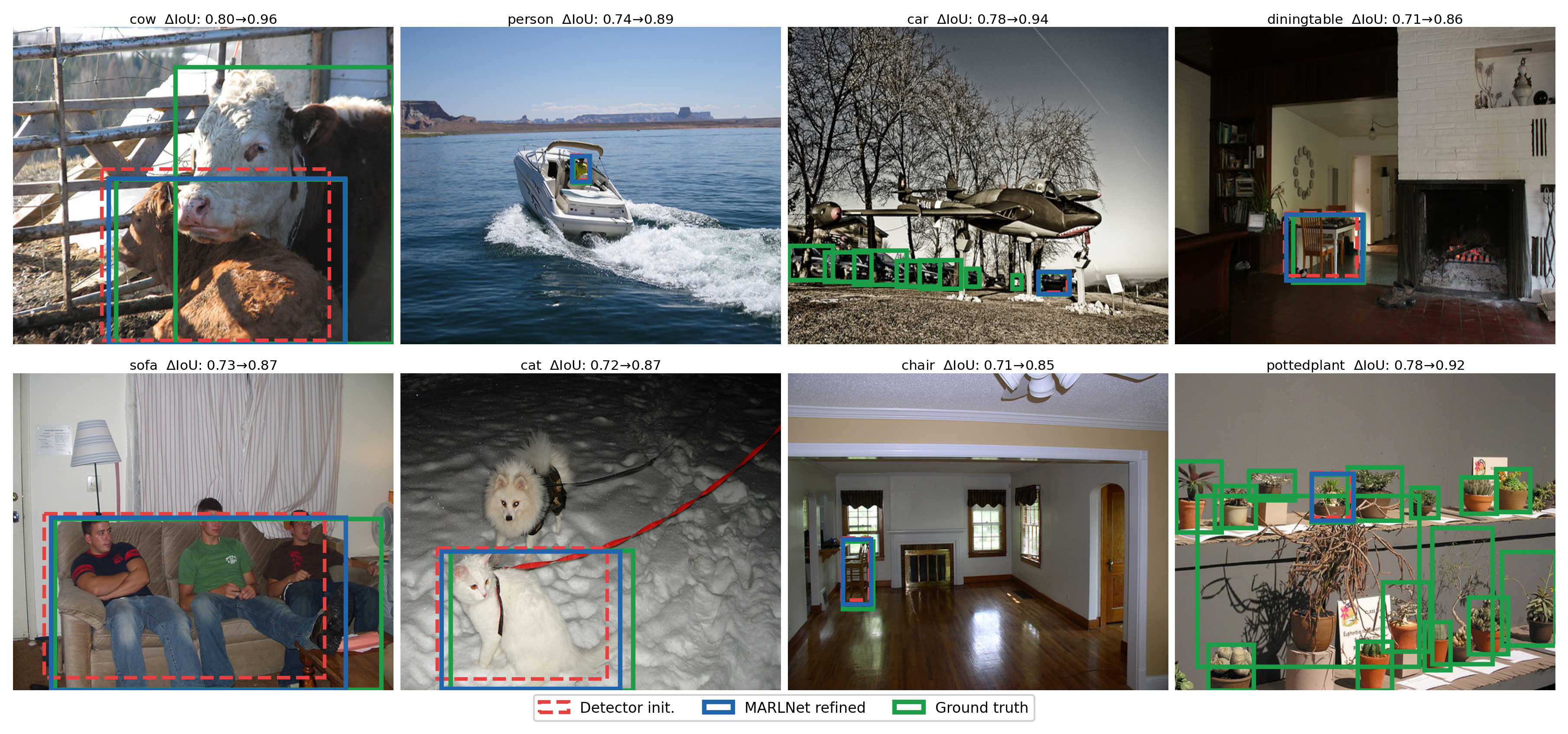}
    \caption{Illustration of refinement process by MARLNet on random images from the VOC 
    validation split.  Each row shows the initial detector box (red) and the final refined box 
    (blue) for a single image.  The agent learns to make small, precise adjustments that often 
    improve IoU with the ground-truth annotation (green).  The smoothness penalty encourages 
    consistent refinement trajectories, as evidenced by the relatively small step sizes and 
    coherent directional movement from red to blue.}
    \label{refinedImages}
\end{figure*}

\begin{table*}[!ht]
  \centering
  \begin{tabular}{lccccccc}
    \toprule
    $\lambda$ & mAP & AP@.5 & AP@.75 & AR & SR@.5 & SR@.7 & $\Delta$IoU \\
    \midrule
    Detector (ref.) & 0.490 & 0.748 & 0.544 & 0.593 & 0.504 & 0.389 & --- \\
    \midrule
    0.05 & 0.428 & 0.745 & 0.477 & 0.534 & 0.506$^\uparrow$ & 0.373 & $-$0.014 \\
    0.10 & \textbf{0.437} & \textbf{0.750} & \textbf{0.491}
         & \textbf{0.544} & \textbf{0.515}$^\uparrow$ & \textbf{0.383} & $-$0.008 \\
    0.20 & 0.434 & 0.748 & 0.483 & 0.539 & 0.511$^\uparrow$ & 0.377 & $-$0.010 \\
    0.50 & 0.408 & 0.741 & 0.432 & 0.512 & 0.499 & 0.360 & $-$0.023 \\
    0.70 & 0.422 & 0.743 & 0.465 & 0.527 & 0.503 & 0.367 & $-$0.018 \\
    \bottomrule
  \end{tabular}
  \caption{Effect of the motion regularization coefficient $\lambda$
    on Pascal VOC 2012 val.
    All variants use the epoch-70 checkpoint.
    The detector baseline is included for reference.
    $\uparrow$ denotes improvement over the detector on that metric. AP@0.5 is stable across the full range, while AP@0.75
    exhibits a non-monotonic pattern with a performance dip
    at intermediate values}
  \label{tab:lambda_ablation}
\end{table*}

\subsection{Results on VisDrone}
\label{sec:visdrone}

Table~\ref{tab:visdrone_results} reports results on the VisDrone
2019 validation split and further per-category results are presented in Table~\ref{tab:visdrone_comparison}.  VisDrone presents a substantially harder
refinement problem than VOC: objects are smaller (median bounding
box area $<$~0.3\% of image area), more densely packed (median
88~annotations per image), and viewed from an aerial perspective
with significant clutter.  The fine-tuned Faster R-CNN achieves
mAP~0.189 on this split, reflecting the inherent difficulty of
the aerial domain.

The method ranking on VisDrone is identical to VOC: the detector
baseline leads on mAP, PPO is the best-performing
refinement method, MARLNet follows, and the heuristic degrades
most severely.  This cross-dataset consistency in the ordering of
methods is the strongest empirical signal of the paper -- the
finding holds across two datasets that differ substantially in
image resolution, viewpoint, object scale, and domain. PPO achieves SR@.5~$=$~0.694 versus the detector's 0.669,
an improvement of $+$0.025. 

This gain is consistent with the relative weakness of the base
detector on this domain (mAP~0.189 vs 0.490 on VOC): when initial
proposals are substantially misaligned, unconstrained policy updates
have greater scope to improve SR@.5 before encountering the
representational ceiling.
On VOC, by contrast, plain PPO does not improve SR@.5 over the
detector ($0.501$ vs $0.504$); it is MARLNet with motion
regularization that achieves the gain ($+0.011$ at $\lambda{=}0.10$).
MARLNet ($\lambda{=}0.05$) achieves SR@.5~$=$~0.666 on VisDrone,
slightly below the detector's 0.669, which we attribute to the
smoothness penalty constraining the step size available to correct
the larger initial misalignments prevalent in the aerial domain.

The lambda ablation on VisDrone (Table~\ref{visdrone_lambda_ablation})
replicates the non-monotonic pattern observed on VOC: intermediate
values ($\lambda \in \{0.1, 0.5\}$) underperform both the low
($\lambda = 0.05$) and high ($\lambda = 0.70$) settings.
$\lambda = 0.70$ achieves the best mAP (0.175) and AP@0.75
(0.170) on VisDrone, consistent with the VOC finding that stronger
smoothness constraints encourage earlier stopping and better
generalization.  The replication of this pattern across both
datasets strengthens the precision-generalization tradeoff
interpretation established in Section~\ref{sec:ablation}.

\begin{table*}[!ht]
  \centering
  \begin{tabular}{lcccccccc}
    \toprule
    \textbf{Method} &
    \textbf{mAP} & \textbf{AP@.5} & \textbf{AP@.75} & \textbf{AR} &
    \textbf{mIoU} & \textbf{SR@.5} & \textbf{SR@.7} &
    \textbf{$\Delta$IoU} \\
    \midrule
    Detector Only (init.)   & 0.189 & 0.337 & 0.187 & 0.280
                            & 0.566 & 0.669 & 0.439 & --- \\
    \midrule
    Heuristic        & 0.140 & 0.257 & 0.134 & 0.266
                            & 0.559 & 0.671 & 0.418 & $-$0.007 \\
    PPO         & \textbf{0.177} & \textbf{0.339}
                            & \textbf{0.172} & \textbf{0.267}
                            & 0.547 & \textbf{0.694} & \textbf{0.423}
                            & $-$0.020 \\
    MARLNet ($\lambda = 0.05$)          & 0.167 & 0.333 & 0.156 & 0.255
                            & \textbf{0.551} & 0.666 & 0.414 & $-$0.015 \\
    \bottomrule
  \end{tabular}
  \caption{Localization results on the VisDrone 2019 validation split
    (548 images). Initial bounding-box hypotheses are provided by a
    Faster R-CNN detector fine-tuned on VisDrone training data.
    \textbf{Bold} indicates the best value among refinement methods.}
  \label{tab:visdrone_results}
\end{table*}

\begin{figure*}[!ht]
    \centering
    \includegraphics[width=0.90\textwidth]{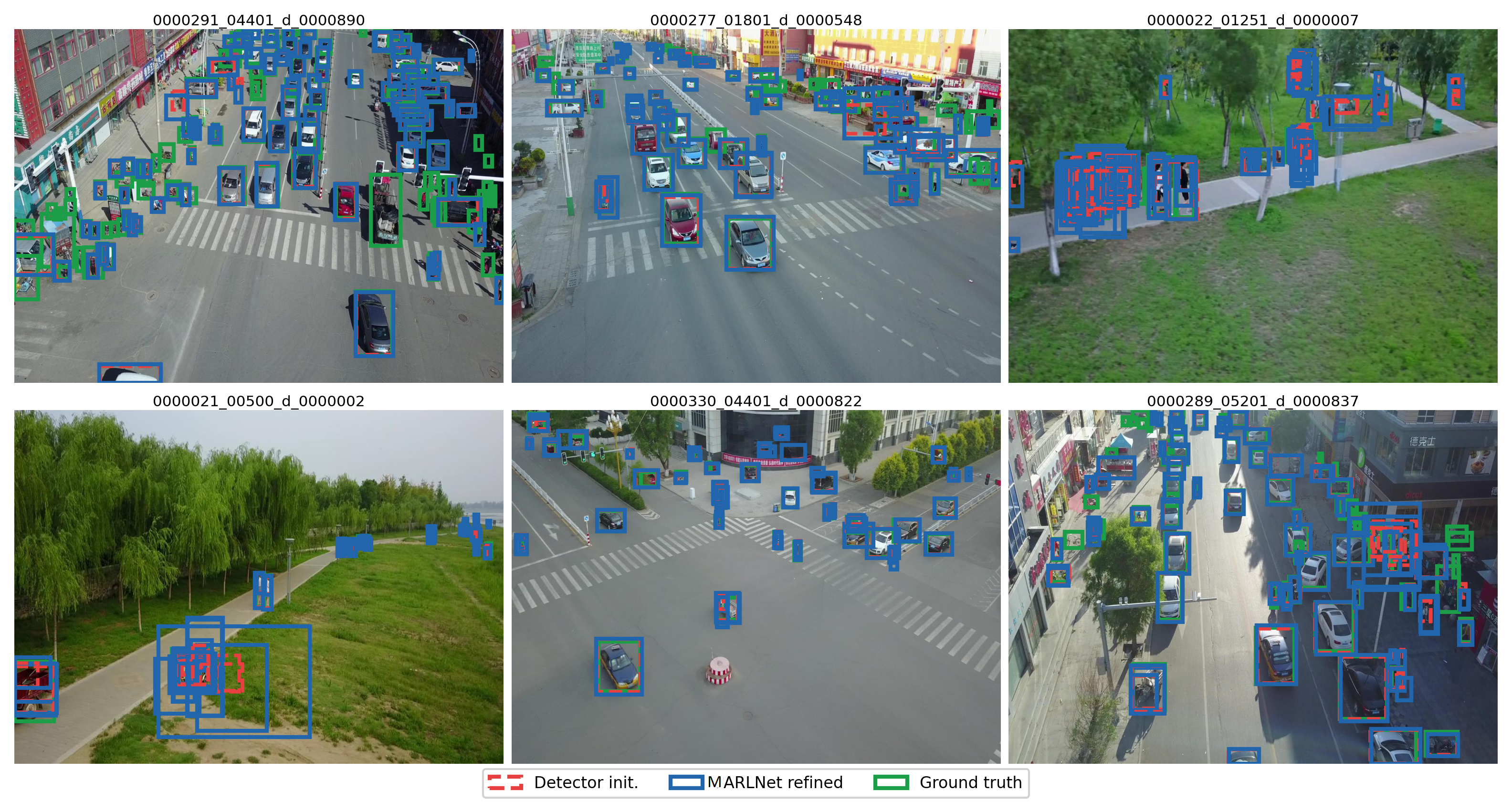}
    \caption{Illustration of refinement process by MARLNet on random images from the VisDrone 2019 validation split.  
    Each group of three columns shows the initial detector box (red), the final refined box (blue), and the ground-truth 
    box (green).  MARLNet successfully refines some marginally-aligned boxes over the 0.5 IoU threshold, but does not 
    systematically correct localization errors, consistent with the feature-information ceiling analysis.}
    \label{refinedVisdrone}
\end{figure*}

\begin{table*}[!ht]
\centering
\begin{tabular}{lcccccccc}
\toprule
\textbf{$\lambda_{\text{phys}}$} & \textbf{mAP} & \textbf{AP@0.5} & \textbf{AP@0.75} & \textbf{SR@0.5} & \textbf{SR@0.7} & \textbf{$\Delta$IoU} \\
\midrule
0.05 (reported) & 0.1671 & 0.3333 & 0.1555 & 0.6663 & 0.4141 & -0.0150 \\
0.1  & 0.1434 & 0.3238 & 0.1014  & 0.6613 & 0.3675 & -0.0391 \\
0.2  & 0.1742 & 0.3356 & 0.1631  & 0.6666 & 0.4221 & -0.0114 \\
0.5  & 0.1681 & 0.3330 & 0.1584  & 0.6634 & 0.4127 & -0.0172 \\
0.7  & 0.1751 & 0.3350 & 0.1698 & 0.6761 & 0.4233 & -0.0121 \\
\bottomrule
\end{tabular}
\caption{Ablation results for different smoothness penalty weights $\lambda_{\text{phys}}$ on visDrone 2019 validation split. All models trained for 70 epochs with identical settings. AP@0.5 is stable across the full range, while AP@0.75 exhibits a non-monotonic pattern with a performance dip at intermediate values.}
\label{visdrone_lambda_ablation}
\end{table*}

\subsection{Effect of Motion regularization Strength}
\label{sec:ablation}

Tables~\ref{tab:lambda_ablation} and~\ref{visdrone_lambda_ablation}
report the effect of varying the smoothness coefficient
$\lambda \in \{0.05, 0.10, 0.20, 0.50, 0.70\}$ on VOC and
VisDrone respectively. Across both datasets, the relationship between $\lambda$ and
performance is non-monotonic, but the specific patterns differ
in ways that illuminate how detector quality and object scale interact with the smoothness prior.

\paragraph{VOC.}
On VOC, $\lambda{=}0.10$ achieves the best overall performance:
mAP~0.437, AP@.5~0.750, and SR@.5~0.515, the highest SR@.5
in the experiment and a gain of $+0.011$ over the base detector.
$\lambda{=}0.20$ follows closely (mAP~0.434, SR@.5~0.511),
also exceeding the detector on SR@.5.
Performance degrades at $\lambda{=}0.50$ (mAP~0.408,
SR@.5~0.499, $\Delta$IoU~$-0.023$), the weakest configuration
on VOC, where heavy smoothing causes the policy to commit
prematurely before meaningful refinement.
$\lambda{=}0.70$ partially recovers (mAP~0.422), suggesting that
very high regularization acts as a conservative early-stopping
mechanism that avoids the worst-case overshooting of intermediate
values.
SR@.5 improvement over the detector on VOC requires motion
regularization: $\lambda \in \{0.05, 0.10, 0.20\}$ all exceed
the detector on this metric, while plain PPO and larger $\lambda$
values do not.

\paragraph{VisDrone.}
On VisDrone, $\lambda{=}0.70$ achieves the best mAP~(0.175) and
AP@.75~(0.170), with $\lambda{=}0.20$ close behind
(mAP~0.174, AP@.75~0.163).
Notably, $\lambda{=}0.10$ is the clear outlier, producing the
worst results across all metrics: mAP~0.143, AP@.75~0.101, and
$\Delta$IoU~$-0.039$ -- by far the most disruptive configuration
across either dataset.
The remaining three values ($\lambda \in \{0.05, 0.50, 0.70\}$)
cluster closely on mAP (0.167--0.175), with higher $\lambda$
generally preferred.
AP@.5 is stable across the full $\lambda$ range on VisDrone
(0.324--0.336, spread 0.012), while AP@.75 varies substantially
(0.101--0.170, spread 0.069), indicating that the smoothness
coefficient primarily affects precision under strict localization
thresholds rather than coarse detection recall.

\paragraph{Cross-dataset comparison.}
From the combined ablation, the most notable finding is
the reversal of $\lambda{=}0.10$ across datasets: it is the best configuration
on VOC and the worst on VisDrone by a significant margin.
This reversal is consistent with the difference in starting
proposal quality between the two domains.
On VOC, where the detector is strong (mAP~0.490) and initial
proposals are well-localized, a moderate smoothness coefficient
allows the policy to make precise, directionally consistent
adjustments without overshooting.
On VisDrone, where the detector is substantially weaker
(mAP~0.189) and proposals require larger corrections, $\lambda{=}0.10$
appears insufficiently constrained to prevent erratic refinement
trajectories -- reflected in its $\Delta$IoU of $-0.039$ --
while $\lambda \geq 0.20$ provides enough regularization to
stabilize the policy.
The practical implication is that $\lambda$ should be treated
as a dataset-dependent hyperparameter calibrated against the
strength of the base detector: a weaker detector benefits from
stronger regularization, while a stronger detector benefits
from lighter regularization that preserves policy expressiveness
for fine-grained adjustments.
\subsection{The Feature Information Ceiling}
\label{sec:ceiling_discussion}

The consistent failure to improve mAP over the detector baseline
on both datasets is not a failure of the RL formulation or the
motion prior -- it is an information-theoretic property of the
problem setup.  The refinement agent observes a crop of the image
centred on the current bounding box, processed by the same
EfficientNet-B0 backbone that informed the detector's initial
prediction.  It therefore operates on a representational basis
that contains no information the detector did not already have
access to.

This ceiling is particularly pronounced on VisDrone, where the
median object occupies a region of approximately $30 \times 20$
pixels in the original image.  When this region is cropped and
upsampled to $224 \times 224$ for EfficientNet-B0, the resulting
features are near-uniform due to the absence of texture variation
at the original scale.  The agent is, in effect, navigating with
near-zero visual signal.  The SR@0.5 improvements suggest that
even under these conditions the policy learns something useful --
but the improvements are bounded by the information available
in the crop.

Closing this gap would require either richer state observations
(multi-scale feature pyramids, global image context, or
appearance models adapted to the target domain) or a weaker
detector initialization that leaves more residual signal for the
agent to exploit.

To verify that the performance ceiling is not an artefact of the
crop-only observation design, we extended the state vector to
524 dimensions by concatenating a full-image EfficientNet-B0
embedding alongside the local crop embedding, providing the agent
with explicit spatial context about the crop's position within
the scene.
Table~\ref{tab:global_local} reports results across four $\lambda$
values. Global$+$local does not consistently outperform crop-only: at
$\lambda \in \{0.05, 0.10\}$, performance drops substantially
-- by $0.078$ and $0.075$ mAP respectively -- while
$\lambda{=}0.20$ produces results nearly identical to the
crop-only variant ($-0.002$ mAP).
The single configuration where global$+$local wins is
$\lambda{=}0.50$ ($+0.008$ mAP), and this gain should be
interpreted cautiously given the instability exhibited at
smaller $\lambda$ values. The sensitivity of global$+$local performance to $\lambda$
is substantially higher than that of the crop-only variant:
the mAP range across tested $\lambda$ values is $0.082$ for
global$+$local versus $0.029$ for crop-only -- a factor of
nearly three -- consistent with a harder optimization
landscape introduced by the higher-dimensional observation.
Taken together, these results confirm that the performance
ceiling arises from the shared representational basis between
the refinement agent and the base detector, rather than from
the limited spatial context of the crop-only observation.
Addressing this ceiling requires architectural changes that
provide the agent access to representations distinct from
those used by the detector -- for instance, a separately
trained feature extractor or a task-specific representation
learned end-to-end with the policy.

\begin{table*}[!htbp]
  \centering
  \begin{tabular}{llcccccc}
    \toprule
    $\lambda$ & Obs. & mAP & AP@.5 & AP@.75 & AR & SR@.5 & $\Delta$IoU \\
    \midrule
    \multirow{2}{*}{0.05}
      & Crop only    & 0.428 & 0.745 & 0.477 & 0.534 & 0.506 & $-$0.014 \\
      & Global+local & 0.350 & 0.617 & 0.380 & 0.476 & 0.477 & $-$0.030 \\
    \midrule
    \multirow{2}{*}{0.10}
      & Crop only    & 0.437 & 0.750 & 0.491 & 0.544 & 0.515 & $-$0.008 \\
      & Global+local & 0.362 & 0.640 & 0.397 & 0.487 & 0.480 & $-$0.030 \\
    \midrule
    \multirow{2}{*}{0.20}
      & Crop only    & 0.434 & 0.748 & 0.483 & 0.539 & 0.511 & $-$0.010 \\
      & Global+local & 0.432 & 0.748 & 0.480 & 0.538 & 0.508 & $-$0.012 \\
    \midrule
    \multirow{2}{*}{0.50}
      & Crop only    & 0.408 & 0.741 & 0.432 & 0.512 & 0.499 & $-$0.023 \\
      & Global+local$^\dagger$         & \textbf{0.416} & \textbf{0.745}
                                       & \textbf{0.448} & \textbf{0.521} & \textbf{0.507}
                                       & $-$0.017 \\
    \midrule
    \multirow{2}{*}{0.70}
      & Crop only    & 0.422 & 0.743 & 0.465 & 0.527 & 0.503 & $-$0.018 \\
      & Global+local & 0.406 & 0.737 & 0.433 &0.508 & 0.492 & $-$0.028 \\
    \bottomrule
  \end{tabular}
  \caption{Crop-only vs.\ global$+$local observation on Pascal
    VOC 2012 val. Adding full-image context (524-dim state)
    does not consistently improve performance, confirming that
    the performance ceiling is representational rather than
    observational. Crop-only results are taken from
    Table~\ref{tab:lambda_ablation}. $\dagger$~denotes the
    single configuration where global$+$local exceeds crop-only
    on mAP.}
  \label{tab:global_local}
\end{table*}

\section{Limitations and Future Work}
\label{sec:future}

Limitations of the current work suggest concrete directions
for future investigation.

\paragraph{Domain-adapted feature extraction.}
EfficientNet-B0 is pre-trained on ImageNet and frozen throughout
RL training.  For aerial imagery such as VisDrone, the
distribution mismatch between ImageNet features and drone-view
objects is severe.  Fine-tuning the feature extractor jointly
with the RL policy -- with careful stabilization to prevent
non-stationarity -- could yield substantially richer crop
representations and break the current information ceiling.

\paragraph{Learned motion prior.}
The constant-velocity model in the observation vector is a
hand-designed, domain-agnostic prior.  A natural extension is
to replace it with a learned dynamics model $f_\psi$ that
predicts expected bbox motion conditioned on visual features and
trajectory history.  Such a model could adapt the motion prior
to different motion patterns--pedestrians, vehicles, small
aerial objects--rather than imposing the same kinematic
assumption across all categories.  The smoothness penalty
framework developed here provides a stable training foundation
on which such an extension could be built. We outline the mathematical formulation of this extension in the Appendix~\ref{sec:learned_prior}.

\section{Conclusion}
\label{sec:conclusion}

We present MARLNet, a motion-aware reinforcement learning
framework for iterative bounding-box refinement that integrates
a constant-velocity motion prior into the observation state and
an action smoothness penalty into the reward.
Evaluated on Pascal VOC 2012 and VisDrone 2019, MARLNet trains
stably across all tested regularization strengths and achieves
gains in detection success rate at $\text{IoU} \geq 0.5$ over
the base detector ($+0.011$ on VOC at
$\lambda_\text{phys}{=}0.10$; $+0.007$ on VisDrone at
$\lambda_\text{phys}{=}0.70$), with plain PPO achieving a
larger SR@.5 gain on VisDrone ($+0.025$) where the weaker base
detector leaves more room for unconstrained refinement. The policy converged to a near-one-shot refinement strategy,
making its computational footprint comparable to single-pass
regression methods while retaining the flexibility of
sequential decision-making.

Through systematic reward design analysis, we identified a
reward interference in which combining a constant-velocity
deviation penalty with an absolute IoU term causes trigger
collapse by epoch~26 -- reducing the agent to maximum-step
jitter and cutting mAP to 0.035 from a detector baseline of
0.490. Replacing the kinematic deviation penalty with an action
smoothness penalty eliminates this failure entirely.
This formulation encodes the correct inductive bias for
static-image refinement -- trajectories should be smooth,
not perpetually accelerating -- and produces stable training
across a wide range of penalty weights on both datasets.

A consistent finding across both datasets is that no
refinement method surpasses the detector on mAP.
We traced this to a representational ceiling: the agent
operates on crop features from the same EfficientNet-B0
backbone as the detector and therefore cannot access signal
the initialisation did not already encode. A global-plus-local observation ablation confirmed the
ceiling is representational rather than observational --
providing a concrete architectural diagnosis for future
work rather than a fundamental limitation of the RL
approach itself.

These findings offer practical guidance for building
motion-regularized RL refinement systems: the smoothness
penalty is a stable and effective motion prior, the choice
of regularization strength should be calibrated to base
detector quality, and breaking the performance ceiling
will require feature representations that go beyond the
detector's own backbone.

{
    \small
    \bibliographystyle{ieeenat_fullname}
    \bibliography{main}
}

\clearpage
\setcounter{page}{1}
\maketitlesupplementary

\section{Implementation Details}
\label{sec:impl}
\paragraph{Network architecture and feature extraction.}
The shared MLP backbone comprises two fully connected layers
of dimension 256 with ReLU activations. The actor head outputs
a 4-dimensional mean vector $\mu_t$, a globally learnable
log-standard-deviation $\log\sigma \in \mathbb{R}^4$, and a
scalar stopping logit; the critic head outputs a scalar value
estimate (approximately $200\,\text{K}$ trainable parameters
total). EfficientNet-B0, pre-trained on ImageNet-1K, projects
its 1280-dimensional global average pooling output to 256
dimensions via a learned linear layer; crops are resized to
$224 \times 224$ before the forward pass.

\paragraph{Training hyperparameters.}
Table~\ref{tab:hyperparams} summarizes all hyperparameters used
in our experiments.

\begin{table}[h]
  \centering
  \caption{MARLNet training hyperparameters.}
  \label{tab:hyperparams}
  \begin{tabular}{lcc}
    \toprule
    \textbf{Parameter} & \textbf{Symbol} & \textbf{Value} \\
    \midrule
    \multicolumn{3}{l}{\textit{PPO}} \\
    Clipping parameter      & $\epsilon$               & $0.2$  \\
    Discount factor         & $\gamma$                 & $0.99$ \\
    GAE parameter           & $\lambda_{\mathrm{GAE}}$ & $0.95$ \\
    Steps per epoch         & $N$                      & $2048$ \\
    PPO update epochs       & $K$                      & $10$   \\
    Minibatch size          & $M$                      & $128$  \\
    \midrule
    \multicolumn{3}{l}{\textit{Loss coefficients}} \\
    Value loss coefficient  & $c_v$                    & $0.5$  \\
    Entropy coefficient     & $c_e$                    & $0.02$ \\
    \midrule
    \multicolumn{3}{l}{\textit{Reward function}} \\
    IoU improvement weight  & $\alpha$                 & $5.0$  \\
    Step cost               & $c_{\mathrm{step}}$      & $0.05$ \\
    Smoothness weight       & $\lambda_{\mathrm{phys}}$& $0.05$ \\
    Terminal bonus scale    & $C_{\mathrm{bonus}}$     & $5.0$ \\
    Stopping threshold      & $\tau_{\mathrm{stop}}$   & $0.6$  \\
    \midrule
    \multicolumn{3}{l}{\textit{Episode and optimization}} \\
    Max refinement steps    & $T_{\max}$               & $15$   \\
    Max action delta        & $\delta_{\max}$          & $0.05$ \\
    Learning rate           & --                      & $2 \times 10^{-4}$ \\
    Optimizer               & --                      & Adam~\cite{kingma2017adam} \\
    Gradient clip norm      & --                      & $0.5$  \\
    \bottomrule
  \end{tabular}
\end{table}

\paragraph{Computational resources.}
All experiments are conducted on a single NVIDIA GeForce RTX~5060
GPU with 8~GB of VRAM and 64~GB of system RAM on 13th Gen Intel Core i7--13700F with clock frequency of 2.10 GHz. 
Training for 70 epochs on the Pascal VOC2012 and VisDrone2019-DET training split takes
approximately 3 to 4 hours.

\section{More Results on VOC \& VisDrone}
Tables~\ref{tab:voc_comparison} and~\ref{tab:visdrone_comparison}
report per-category AP@0.5 for all methods on VOC and VisDrone
respectively, complementing the aggregate metrics in the main paper.

\begin{table*}[!htbp]
\centering
\small 
\begin{tabular}{lcccc}
\toprule
\textbf{Category} & \textbf{Detector only} & \textbf{Heuristic} & \textbf{PPO} & \textbf{MARLNet} \\
\midrule
aeroplane   & 0.8210 & 0.7806 & \textbf{0.8283} & 0.8270 \\
bicycle     & \textbf{0.7818} & 0.6615 & 0.7736          & 0.7817 \\
bird        & \textbf{0.7726} & 0.7084 & 0.7651          & 0.7588 \\
boat        & 0.5814          & 0.5174 & \textbf{0.5838} & 0.5689 \\
bottle      & \textbf{0.6545} & 0.6079 & 0.6484          & 0.6519 \\
bus         & 0.8444          & 0.8120 & \textbf{0.8452} & 0.8444 \\
car         & \textbf{0.7041} & 0.6321 & 0.6992          & 0.6978 \\
cat         & \textbf{0.8838} & 0.8041 & 0.8773          & 0.8751 \\
chair       & \textbf{0.5777} & 0.4541 & 0.5767          & 0.5660 \\
cow         & \textbf{0.7970} & 0.6959 & 0.7934          & 0.7849 \\
diningtable & \textbf{0.5360} & 0.4051 & 0.5278          & 0.5275 \\
dog         & \textbf{0.8489} & 0.7417 & 0.8377          & 0.8401 \\
horse       & \textbf{0.8563} & 0.7808 & 0.8508          & 0.8548 \\
motorbike   & \textbf{0.8777} & 0.7825 & 0.8746          & 0.8745 \\
person      & \textbf{0.8642} & 0.7885 & 0.8607          & 0.8607 \\
pottedplant & 0.5755          & 0.4964 & \textbf{0.5793} & 0.5649 \\
sheep       & \textbf{0.7844} & 0.7229 & 0.7803          & 0.7842 \\
sofa        & \textbf{0.6267} & 0.3375 & 0.6215          & 0.6199 \\
train       & 0.8413          & 0.8020 & \textbf{0.8437} & 0.8406 \\
tvmonitor   & \textbf{0.7319} & 0.6879 & 0.7309          & 0.7243 \\
\midrule
\textbf{Mean AP@0.5} & \textbf{0.7481} & 0.6610 & 0.7449 & 0.7424 \\
\bottomrule
\end{tabular}
\caption{Per-category and mean Average Precision (AP@0.5) comparison across different architectures on PASCAL VOC Dataset.}
\label{tab:voc_comparison}
\end{table*}

\begin{table*}[!htbp]
\centering
\small 
\begin{tabular}{lcccc}
\toprule
\textbf{Category} & \textbf{Detector only} & \textbf{Heuristic} & \textbf{PPO} & \textbf{MARLNet} \\
\midrule
pedestrian      & 0.4133          & 0.3634 & \textbf{0.4203} & 0.4087 \\
people          & \textbf{0.3034} & 0.2011 & 0.3009          & 0.2955 \\
bicycle         & \textbf{0.0964} & 0.0420 & 0.0953          & 0.0921 \\
car             & \textbf{0.7482} & 0.7190 & \textbf{0.7482} & 0.7471 \\
van             & \textbf{0.3882} & 0.2102 & 0.3866          & 0.3826 \\
truck           & 0.3365          & 0.2560 & \textbf{0.3384} & 0.3343 \\
tricycle        & 0.2118          & 0.1168 & \textbf{0.2133} & 0.2109 \\
awning-tricycle & 0.0901          & 0.0450 & \textbf{0.0906} & 0.0905 \\
bus             & 0.3805          & 0.2713 & \textbf{0.3851} & 0.3805 \\
motor           & 0.4024          & 0.3494 & \textbf{0.4063} & 0.3907 \\
\midrule
\textbf{Mean AP@0.5} & 0.3371          & 0.2574 & \textbf{0.3385} & 0.3333 \\
\bottomrule
\end{tabular}
\caption{Per-category and mean Average Precision (AP@0.5) comparison across architectures for the VisDrone-2019-DET dataset.}
\label{tab:visdrone_comparison}
\end{table*}

\section{Mathematical Formulation of the Learnable Motion Model}
\label{sec:learned_prior}
While the fixed constant–velocity assumption provides a simple and effective motion prior,
it cannot capture complex object dynamics or camera motion.
To address this limitation, we extend the framework by introducing a
\emph{learnable motion module} that predicts the next bounding box in a data-driven manner.

\paragraph{Parametric Dynamics Model.}
Instead of the handcrafted update
$\hat{b}_{phys,t} = b_t + (b_t - b_{t-1})$,
we define a differentiable mapping
\[
\hat{b}_{phys,t} = f_\psi(b_t, \Delta b_t, f_t),
\]
where $f_\psi(\cdot)$ is a neural network with parameters~$\psi$.
The inputs consist of the current box $b_t=(x_t, y_t, w_t, h_t)$,
its motion $\Delta b_t=b_t-b_{t-1}$,
and the visual feature embedding $f_t$ extracted from the cropped region.
The network outputs a motion offset that estimates the next physically plausible box.

\paragraph{Motion Consistency Loss.}
During training, the predicted box $\hat{b}_{phys,t}$ is encouraged to match the
agent’s actual next box $b_{t+1}$ through a self-consistency term:
\[
\mathcal{L}_{phys} = \|\,b_{t+1} - f_\psi(b_t, \Delta b_t, f_t)\,\|_2^2.
\]
Optionally, when frame-wise ground truth $b^*_{t+1}$ is available,
a supervised variant
$\mathcal{L}_{phys}^{GT} = \|\,b^*_{t+1} - f_\psi(b_t, \Delta b_t, f_t)\,\|_2^2$
can be used to further guide learning.

\paragraph{Joint Optimization.}
The motion model is trained jointly with the PPO policy~$\pi_\theta$.
At each step, the reward incorporates the learned prediction:
\[
r_t = \alpha\,\Delta\text{IoU}
      - \lambda_{phys}\,\|\,b_t - \hat{b}_{phys,t}\,\|_2,
\]
and the overall objective becomes
\[
\max_{\theta,\psi}
\;
\mathbb{E}_t
  \big[ J_{PPO}(\theta; r_t(\psi)) \big]
  - \eta\,\mathcal{L}_{phys},
\]
where $\eta$ balances reinforcement and motion-consistency terms.
Because $r_t$ depends on the differentiable function~$f_\psi$,
gradients can propagate through the motion module, enabling end-to-end learning.

\paragraph{Interpretation.}
The learnable motion model acts as a data-driven dynamics prior that
adapts to diverse motion patterns.
While the PPO agent optimizes localization accuracy via IoU-based reward,
$f_\psi$ learns to model how bounding boxes evolve across time.
Their co-training yields policies that are both accurate and physically coherent,
bridging model-free and model-based reinforcement learning within a unified framework.

\section{Heuristic Refinement Policy}
\label{app:heuristic}

We describe the greedy detector-rescore heuristic used as a
non-learned baseline throughout the experiments.
The method requires no training and no ground-truth supervision
at inference time.
It shares the same base detector as MARLNet, ensuring a
fair comparison: both methods operate from identical initial
proposals and use the same Faster R-CNN backbone.

\paragraph{Design rationale.}
The heuristic is motivated by a simple observation: if a
candidate box better encloses an object, the detector should
assign higher confidence to a proposal that overlaps it.
Rather than learning a policy to maximise IoU directly,
the heuristic uses the detector's own confidence score as
a proxy for localization quality, performing a greedy
coordinate search to find the nearby box that the detector
scores most highly.
This proxy is imperfect -- detector confidence and IoU
with ground truth are not perfectly correlated -- which
explains why the heuristic is competitive but bounded
relative to methods that optimise the true localization
signal directly.

\begin{algorithm}[t]
\caption{Greedy Heuristic Refinement}
\label{alg:heuristic}
\textbf{Input:} Image $I$; initial detection $(b_0, s_0)$;
detector $\mathcal{D}$; max steps $T{=}8$;
scale factors $\mathcal{S}{=}\{0.9, 1.0, 1.1\}$;
shift fraction $\delta{=}0.15$;
candidate limit $N_c{=}9$;
overlap threshold $\tau{=}0.5$;
improvement threshold $\varepsilon{=}10^{-4}$ \\
\textbf{Output:} Refined box $b^*$

\begin{algorithmic}[1]

\Function{GenerateCandidates}{$b,\,\mathcal{S},\,\delta,\,N_c$}
  \State Extract centre $(c_x, c_y)$ and size $(w, h)$ from $b$
  \State $\text{Shifts} \gets
    \{(0,0),\;(\delta,0),\;(-\delta,0),\;(0,\delta),\;(0,-\delta)\}$
  \State $\mathcal{C} \gets \emptyset$
  \For{$\sigma \in \mathcal{S}$}
    \For{$(d_x, d_y) \in \text{Shifts}$}
      \State Construct box centred at
        $(c_x + d_x w,\; c_y + d_y h)$
        with size $(\sigma w,\; \sigma h)$
      \State Clamp to image boundaries; add to $\mathcal{C}$
    \EndFor
  \EndFor
  \State \Return first $N_c$ unique boxes from $\mathcal{C}$
\EndFunction

\vspace{4pt}
\State $b \gets b_0,\quad s \gets s_0$

\For{$t = 1$ \textbf{to} $T$}
  \State $\mathcal{C} \gets
    \textsc{GenerateCandidates}(b,\,\mathcal{S},\,\delta,\,N_c)$
  \State $\mathcal{O} \gets \mathcal{D}(I)$
    \hfill\textit{// run detector once per step; reuse for all candidates}
  \For{each candidate $c \in \mathcal{C}$}
    \State $\text{score}(c) \gets
      \max\limits_{\substack{o_i \in \mathcal{O} \\
        \mathrm{IoU}(c,\, o_i) \geq \tau}}
      \mathrm{conf}(o_i)$
      \hfill\textit{// 0 if no $o_i$ overlaps $c$}
  \EndFor
  \State $b' \gets \arg\max_{c \in \mathcal{C}}\;\text{score}(c),
    \quad s' \gets \max_{c \in \mathcal{C}}\;\text{score}(c)$
  \If{$s' \leq s + \varepsilon$}
    \State \textbf{break}
    \hfill\textit{// no improvement; terminate early}
  \EndIf
  \State $b \gets b',\quad s \gets s'$
\EndFor

\State \Return $b^* \gets b$

\end{algorithmic}
\end{algorithm}

\paragraph{Candidate generation and scoring.}
At each step, \textsc{GenerateCandidates} produces up to
$N_c{=}9$ boxes by combining three scale factors
($\times 0.9$, $\times 1.0$, $\times 1.1$) with five centre
offsets (no shift, and $\pm 15\%$ of box dimensions along each
axis), deduplicating the resulting $3{\times}5$ grid.
Each candidate is scored as the maximum detector confidence
among detector boxes with $\mathrm{IoU} \geq \tau{=}0.5$,
and zero when none overlap. The detector runs once per step
and its outputs are reused across all candidates. Refinement
terminates when no candidate improves the score by more than
$\varepsilon{=}10^{-4}$, or when $T{=}8$ steps are exhausted.

\section{Discussion on Results of Experiment}
\label{sec:dis_results}
\begin{figure*}[!ht]
  \centering
  \includegraphics[width=\linewidth]{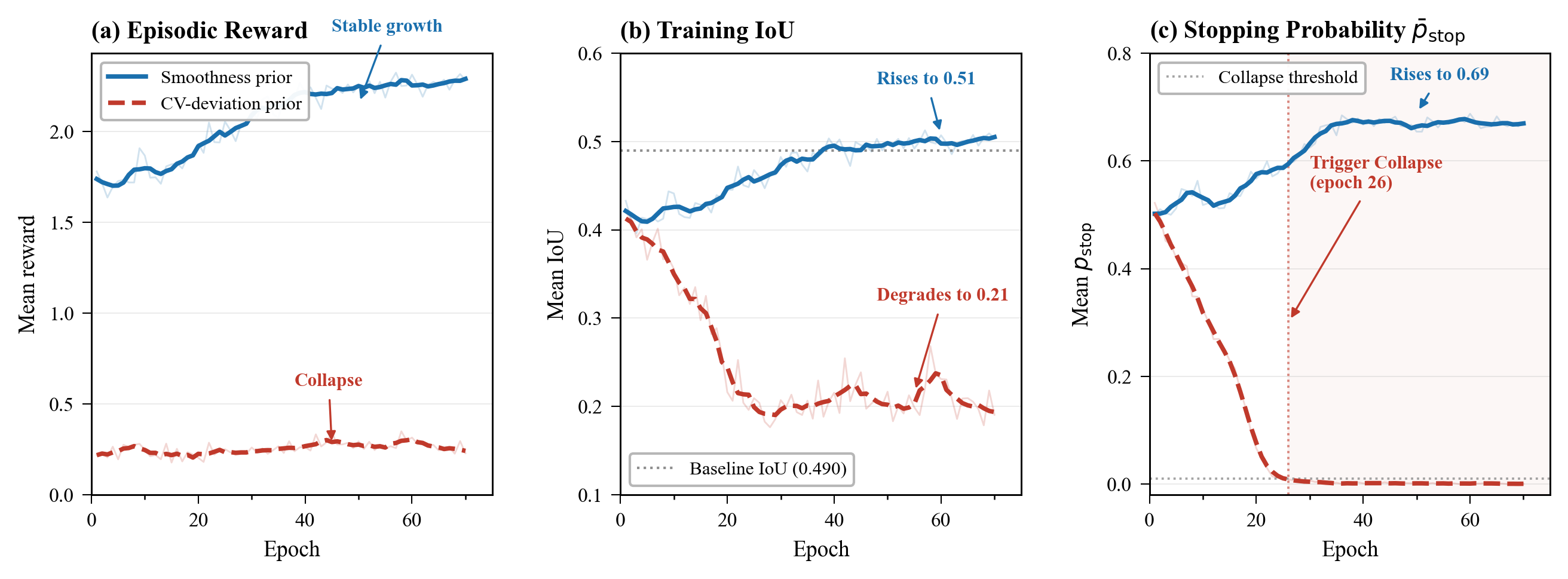}
  \caption{Training dynamics comparing the two motion prior
    formulations over 70 epochs. \emph{(a)} Mean episodic reward:
    the smoothness prior (blue, solid) grows steadily, while the
    CV-deviation prior (red, dashed) remains flat near zero
    throughout training. \emph{(b)} Mean training IoU: the
    smoothness prior rises to 0.51 before the run was stopped at
    epoch 69, consistently approaching the detector initialization
    IoU of 0.487 (dotted line) and exceeding it; the CV-deviation
    prior degrades monotonically from 0.42 to 0.21, falling far
    below the detector baseline. \emph{(c)} Mean stopping
    probability $\bar{p}_\mathrm{stop}$: the smoothness prior
    learns a well-calibrated stopping signal that rises from 0.50
    to 0.69; the CV-deviation prior undergoes trigger collapse by
    epoch 26 (vertical dotted line), after which $\bar{p}_\mathrm{stop}$
    remains at zero for the remainder of training. The shaded red
    region marks the collapse regime. Together, panels (b) and (c)
    confirm that the CV-deviation formulation fails not merely by
    producing lower reward, but by causing the agent to lose its
    stopping mechanism entirely, reducing it to maximum-step
    jitter.}
  \label{fig:training_curves}
\end{figure*}

\subsection{Reward Design Analysis}
\label{sec:ablation}

\paragraph{The CV-penalty reward causes catastrophic failure.}
When the motion penalty is defined as deviation from the
constant-velocity prediction -- $\lambda_\mathrm{phys}
\|b_{t+1} - \hat{b}^\mathrm{cv}_{t+1}\|_2$ -- training
collapses by epoch 26.  The mean stopping probability
$\bar{p}_\mathrm{stop}$ falls below 0.01 and never recovers,
reducing the model to random per-step jitter over all $T_\mathrm{max}$
steps. Figure~\ref{fig:training_curves}(b) shows mean training IoU
degrading from 0.42 to 0.21, confirming complete localization
failure.

The failure mechanism is a reward accumulation
imbalance.  At initialization, each step yields approximately
$\beta \cdot \mathrm{IoU}_t \approx 0.42$ from the absolute
IoU term alone.  Accumulated over $T_\mathrm{max} = 15$ steps,
this exceeds the terminal bonus ($C_\mathrm{bonus} \cdot
\mathrm{IoU} \leq 10 \cdot 0.6 = 6.0$ for typical final IoU)
by a large margin, making it always more profitable for the agent
to run all steps rather than stop.  The policy consequently
learns to suppress $p_\mathrm{stop}$ to zero, eliminating the
only mechanism for obtaining well-timed terminal bonuses.

This failure is compounded by the CV prediction itself:
the constant-velocity model extrapolates continued
acceleration in the box's current direction, whereas a
converging agent must decelerate toward the target.
The CV penalty therefore actively discourages the convergent
behaviour the agent should exhibit.

\paragraph{The smoothness penalty recovers stable training.}
Replacing the CV penalty with the action-smoothness penalty
eliminates trigger collapse: $\bar{p}_\mathrm{stop}$ rises
monotonically from 0.50 to 0.69 over 70 epochs and the agent
learns a meaningful stopping strategy.  The smoothness penalty
makes no claim about where the box should be at the next step;
it only constrains how abruptly the action direction can change,
which is the physically reasonable inductive bias for a
refinement agent operating on static images.

\subsection{Training Dynamics}
\label{sec:training_dynamics}

Figure~\ref{fig:training_curves} plots three training metrics --
mean episodic reward, mean IoU, and mean stopping probability
-- over the course of training for MARLNet and the
CV-penalty variant.

Two observations from the training curves are worth noting.
MARLNet's mean training IoU rises from 0.42 at epoch~1 to 0.51 by epoch~70, confirming that the policy
improves localization over training. The stopping probability
rises monotonically from 0.50 to 0.69, indicating the agent
learns to fire the termination trigger at well-aligned
proposals rather than exhausting the maximum step budget.
Validation performance stabilizes by epoch~60, which is the
checkpoint reported throughout the paper.
\subsection{Inference Efficiency}
\label{sec:latency_discussion}

\begin{figure*}[!ht]
    \centering
    \includegraphics[width=0.9\textwidth]{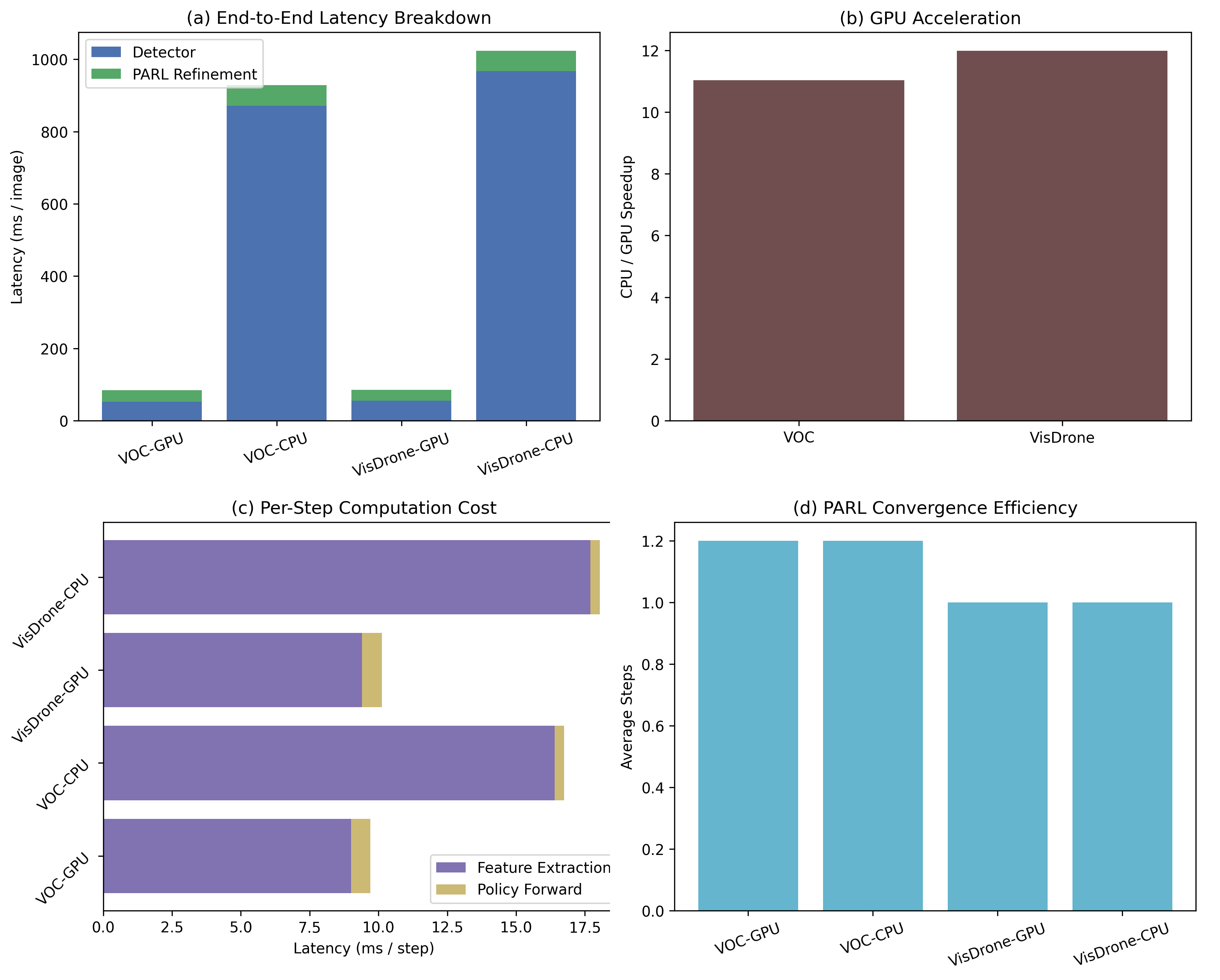}
    \caption{Inference latency of MARLNet on GPU and CPU across both datasets.  The policy forward pass is negligible compared to feature extraction and the detector, confirming that the RL component does not introduce a significant computational bottleneck.}
    \label{fig:latency}
\end{figure*}

Table~\ref{tab:latency} and Figure~\ref{fig:latency} report
inference latency across both datasets and devices.  On GPU,
the full MARLNet pipeline runs in approximately 84--85~ms
end-to-end, corresponding to approximately 12~FPS.  The policy
forward pass accounts for less than 1~ms of this total; the
dominant cost is EfficientNet-B0 crop extraction (9--10~ms per
step).  On CPU, the bottleneck shifts entirely to Faster R-CNN
($\sim$900~ms), with MARLNet adding only 56--57~ms overhead --
6--7\% of total pipeline time.

The average of 1.0--1.2 refinement steps per episode is
a notable finding.  The policy converged to a near-one-shot
strategy through reward shaping rather than architectural
constraint: the terminal bonus makes immediate stopping at a
reasonable IoU preferable to continued refinement in expectation.
This makes MARLNet computationally comparable to single-pass
regression-based refinement methods, while retaining the
theoretical flexibility of iterative sequential decision-making.

\begin{table*}[!ht]
  \centering
  \begin{tabular}{llccccc}
    \toprule
    \textbf{Dataset} & \textbf{Device} &
    \textbf{Steps} &
    \textbf{Feat./step} &
    \textbf{Refinement} &
    \textbf{Detector} &
    \textbf{End-to-end} \\
    \midrule
    VOC      & GPU & 1.2 & 9.0 ms  &  32 ms &  52 ms &  84 ms \\
    VOC      & CPU & 1.2 & 16.4 ms &  57 ms & 871 ms & 929 ms \\
    VisDrone & GPU & 1.0 & 9.4 ms  &  30 ms &  55 ms &  85 ms \\
    VisDrone & CPU & 1.0 & 17.7 ms &  56 ms & 968 ms & 1024 ms \\
    \bottomrule
  \end{tabular}
  \caption{Inference latency of MARLNet (mean ms per image, 200 images
    after 50 warmup iterations). Feature extraction and policy forward
    are reported per step; total refinement is summed across all steps.
    Hardware: NVIDIA GeForce RTX 5060 (GPU) and Intel i7 13th Gen CPU.}
  \label{tab:latency}
\end{table*}

\end{document}